\documentclass{article}
     \PassOptionsToPackage{numbers, compress}{natbib}

     \usepackage[preprint]{neurips_2020}

\usepackage[utf8]{inputenc} 
\usepackage[T1]{fontenc}    
\usepackage{hyperref}       
\usepackage{url} 
\usepackage{booktabs}       
\usepackage{amsfonts}       
\usepackage{nicefrac}       
\usepackage{microtype}      
\usepackage{algorithm}
\usepackage[noend]{algpseudocode}
\usepackage{graphicx}
\usepackage{xcolor}
\usepackage{amsmath}
\usepackage{enumitem}

\title{Iterate \& Cluster: Iterative Semi-Supervised Action Recognition}

\author{%
  Jingyuan Li\\
  Department of Electrical \& Computer Engineering\\
  University of Washington\\
  Seattle, WA, 98195 \\
  \texttt{jingyli6@uw.edu} \\
   \And
   Eli Shlizerman \\
   Department of Applied Mathematics \\
   Department of Electrical \& Computer Engineering \\
   University of Washington \\
   Seattle, WA, 98195\\
   \texttt{shlizee@uw.edu} \\
}

\begin{document}

\maketitle

\begin{abstract}
We propose a novel system for active semi-supervised feature-based action recognition. Given time sequences of features tracked during movements our system clusters the sequences into actions. Our system is based on encoder-decoder unsupervised methods shown to perform clustering by self-organization of their latent representation through the auto-regression task. These methods were tested on human action recognition benchmarks and outperformed non-feature based unsupervised methods and achieved comparable accuracy to skeleton-based supervised methods. However, such methods rely on K-Nearest Neighbours (KNN) associating sequences to actions, and general features with no annotated data would correspond to approximate clusters which could be further enhanced. Our system proposes an iterative semi-supervised method to address this challenge and to actively learn the association of clusters and actions. The method utilizes latent space embedding and clustering of the unsupervised encoder-decoder to guide the selection of sequences to be annotated in each iteration. Each iteration, the selection aims to enhance action recognition accuracy while choosing a small number of sequences for annotation. We test the approach on human skeleton-based action recognition benchmarks assuming that only annotations chosen by our method are available and on mouse movements videos recorded in lab experiments. We show that our system can boost recognition performance with only a small percentage of annotations. The system can be used as an interactive annotation tool to guide labeling efforts for 'in the wild' videos of various objects and actions to reach robust recognition. 
\end{abstract}

\section{Introduction}





\begin{figure}
    \begin{minipage}{0.78\textwidth}
      \centering
        \includegraphics[width=0.95\textwidth]{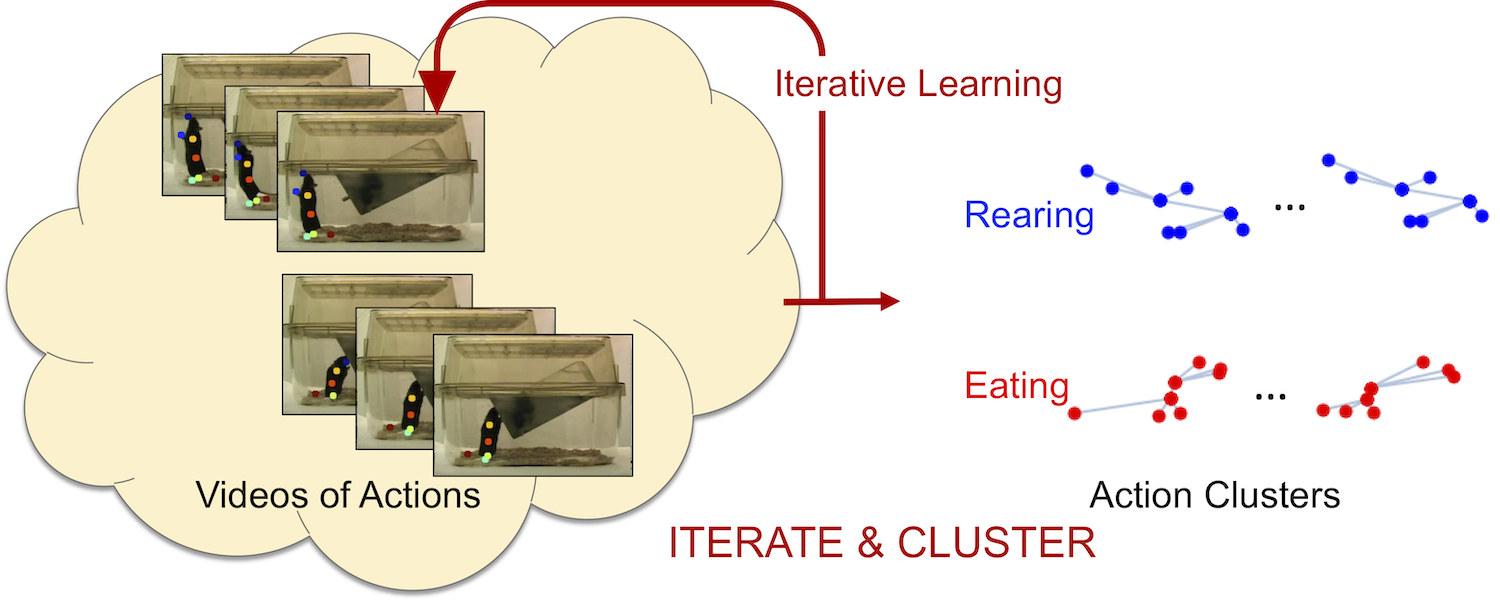}
    \end{minipage}%
    \begin{minipage}{0.22\textwidth}
      \caption{Iterative \& Cluster uses features detected in videos to associate actions through semi-supervised iterative learning.}
    	\label{fig:Teaser}
\end{minipage}
\end{figure}

Videos of objects performing actions include a variety of informational cues such that action recognition system using videos aims to leverage this information to assign label for the action being performed by objects in the video frames sequence. Due to the variety of informational cues in the video frames, there is an assortment of methods focusing on different features and aim to associate them with actions. A direct approach is to attempt to perform action recognition directly from the image frames of the video (RGB) or even to consider image and depth information (RGB+D). It is indeed possible to obtain robust action recognition directly from video, however it is often the case that the pixel data includes redundant features. For example, features from the background of the video are often irrelevant to objects actions. Due to these limitations, to achieve efficient performance, most RGB based methods require an extensive supervised training with limited number of actions and comprehensive annotated datasets.
Complementary to direct consideration of video frames as input, markerless feature detector methods offer a rapid detection of cues in each frame. For common objects, such as people or their faces, these are standard features that define pose estimation such as the skeleton joints or contours, respectively~\cite{rahmani2014hopc, wang2016temporal}. For general objects, e.g., animals, similar methods allow to pre-select and define features. Notable examples are DeepCut~\cite{pishchulin2016deepcut} and DeepLabCut which perform markerless pose estimation from video based on transfer learning~\cite{mathis2018deeplabcut}. In addition to detection, it is also possible to track the same features over frames providing correspondence through time. The use of these predefined features is advantageous for action recognition since they filter the unnecessary information and provide concise sequences from which actions can be extracted in a more direct way. Indeed, visual perceptual studies showed that 2D or 3D skeleton data does contain a clear information about most actions and only a few key points (10-12) are enough to stimulate neural responses to actions~\cite{johansson1973visual}. Moreover, recent works show that the features such as lips contours or skeleton can be even generated from audio only~\cite{suwajanakorn2017synthesizing,shlizerman2018audio}.
Several systems have been introduced for action recognition from body keypoints. The main challenge in these approaches is to identify the spatial and temporal relation of each action and most system solve it by supervised methods which require annotated datasets, which introduces additional challenge of obtaining  annotations. Indeed, annotations challenges are common and it is often up to the interpretation of the annotator to assign a meaningful label for a given sequence, especially in situations where the annotation is of animal movements. 
Recent advances suggest a possibility to overcome the annotation requirement by the implementation of an unsupervised action recognition system from keypoints and show promising performance~\cite{zheng2018unsupervised,su2019predict,suclustering}. The system uses an encoder-decoder approach to regenerate the sequences of keypoints and self-organizes its latent states to cluster actions. A  K-Nearest Neighbors (KNN) classifier is then used to associate clustered points with actions. It turns out that the KNN classifier also requires training and unsupervised methods replacing KNN association to labels do not perform as well as trained KNN. It is therefore advantageous to consider efficient strategies that will resolve the association with a few annotations. This is the goal of our system. 
Inspired by the success of DeepCut and Active Learning approaches~\cite{pishchulin2016deepcut, activ2012}, we introduce an iterative selection of sequences to be annotated to resolve the association. In particular, our system, \textit{Iterate \& Cluster}, introduces a semi-supervised method to actively learn the association of clusters and actions. The method utilizes the latent space embedding and clustering of the unsupervised encoder-decoder methodology to guide the selection of sequences to be annotated in each iteration. As such the method is able to boost performance of an unsupervised action recognition with only a small numeber of annotations and we demonstrate its robustness on several datasets of skeleton-based action recognition and also mouse movements recognition, (Fig.~\ref{fig:Teaser}). 


\section{Related Work}
Skeleton-based action recognition has been studied extensively and a variety of approaches were introduced to organize movement sequences into actions with \textit{supervised learning} which trains the system with an annotated dataset of actions. Earlier studies focused on computing local statistics from both spatial saliences and action motions~\cite{xia2012view, wang2013learning, vemulapalli2014human} and associated them to actions. More recently methods based on deep learning have been proposed. Such systems are structured with whether underlying Recurrent Neural Network (RNN) or Convolutional Neural Network (CNN) architecture. RNN based methods include an end-to-end hierarchical RNN~\cite{du2015hierarchical} and part-aware long short-term memory (P-LSTM) unit~\cite{shahroudy2016ntu} using internal part based memory sub-cells with new gating strategies for skeleton action recognition. CNN approaches propose to transform skeletons to a series of color images which are then taken as an input into a network where features of spatial temporal information are being extracted~\cite{liu2017enhanced}. Furthermore, as an extension, recent approaches proposed to associate graphs with the skeleton joint information to capture the relations of bones between the joints. Such an extension turns the underlying architecture to a Graph Convolutional Network (GCN) and enhances the action recognition performance. Specifically, the method of Spatial-Temporal Graph Convolutional Network (ST-GCN) proposed a graph with spatial edges connecting keypoints in each time step and temporal edges connecting the same keypoint across different time steps~\cite{yan2018spatial}. While able to reach successful accuracy and robustness, these methods are specific to human-skeleton joints keypoints and would need to be redesigned and retrained if to be considered for action recognition of general objects and movements. Since those methods are supervised, to support such steps it would require annotated datasets of actions to be trained upon.

Due to the challenges with annotation and generalization \textit{unsupervised action recognition} approaches have been recently developed. These approaches do not require annotations instead seek to learn a representation to capture the long-term global motion dynamics in skeleton sequences using an encoder-decoder RNN~\cite{zheng2018unsupervised}. Such a setting showed the promise of self-organization of the network to cluster actions, however, also left the question, how representations should be extracted or the network to be designed to perform high-precision recognition. Recently methods for representation learning enhanced the approach and provided network structure to enhance the representation and hence the performance~\cite{su2019predict,suclustering}. The system uses a Gated Recurrent Unit (GRU) based encoder-decoder RNN and focuses on weakening the decoder by fixing its weights or the inputs into it, such that the latent representation of the encoder, in particular the last state of it, will be promoted to represent different actions as different classes. The system then proposed an auto-encoder KNN classifier for the representation to study the performance on extensive benchmarks of skeleton-based recognition such as NW-UCLA~\cite{wang2014cross}, UWA3D~\cite{rahmani2014hopc} and NTU RGB+D~\cite{shahroudy2016ntu} with various number of actions and samples. The outcome performance on these benchmarks showed a performance comparable to supervised methods when is joint with KNN classifier trained on the dataset. With our system, we will replace this classifier with an efficient semi-supervised approach for active learning to boost performance.

\textit{Semi-supervised} learning finds the middle ground between supervised and unsupervised approaches and is commonly used in the field of computer vision and action recognition. It does rely on labeled data points provided during training and at the same time aims to leverage other properties of the model or data to boost performance or minimize the need for large numbers of labeled data points. Semi-supervised learning is typically used as a fine-tuning process where first unsupervised learning is performed and then it is  fine tuned by a supervised learning procedure. For example, video based action recognition used semi-supervised learning to train deep convolutional generative adversarial network using a large video activity dataset without annotated information and then used the representation and labeled data for achieving robust performance~\cite{ahsan2018discrimnet}. While fine-tuning approaches are common, they are typically implemented on the full annotated dataset and do not necessary show benefit when only a fraction of labels is used. Since unsupervised methods do not require labels at all it is warranted to attempt to perform selection of data points iterative to be annotated to require an effective and minimal annotation effort. In \textit{Iterate \& Cluster} we introduce an iterative semi-supervised strategy that leverages the unsupervised clustering and iteratively adds labels of the most promising data points to enhance the unsupervised recognition accuracy. 

\textit{Active Learning (AL)} aims to efficiently select data points for training to guide annotation~\cite{cohn1994improving,atlas1990training}. It is based on computing a metric representing the utility to decide whether samples should be queried for labeling. Several methods for computing such metrics and selecting according to them were introduced, e.g., information theoretical methods, ensemble approaches, and uncertainty based methods~\cite{activ2012}. DeepCut implement active learning based approach for 
system training~ \cite{pishchulin2016deepcut,mathis2018deeplabcut}. The system takes video frames as input and then reconstructs original frames separately for corresponding time steps predicting keypoints positions, along with the probability describing how believable the predicted position for each keypoints. At a later stage, it selects outlier samples for re-annotation and retraining the network. By doing so it enhances performance. In \textit{Iterate \& Cluster} we consider uncertainty based methods which include Variation Ratio~\cite{Variationfreeman} and Maximum Entropy\cite{shannon1948mathematical} to measure the lack of confidence in sequneces and these will combined with the unsupervised clustering and  selected for annotation ~\cite{joshi2009multi,gal2017deep}. 

\section{Methods}\label{Methods}

\begin{figure}[!t]
    \begin{minipage}[l]{.49\textwidth}
        \centering
        \includegraphics[width=1\textwidth]{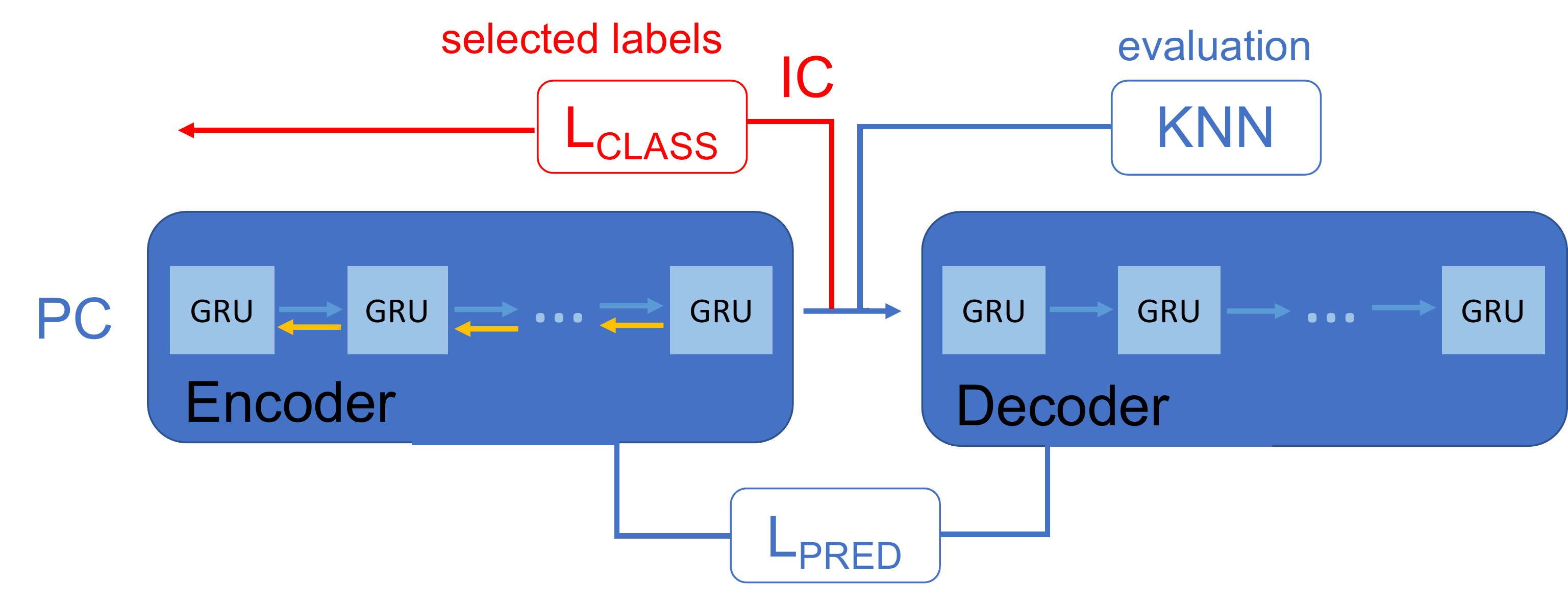}
    \end{minipage}%
    \begin{minipage}[r]{0.49\textwidth}
        \includegraphics[width=1\textwidth]{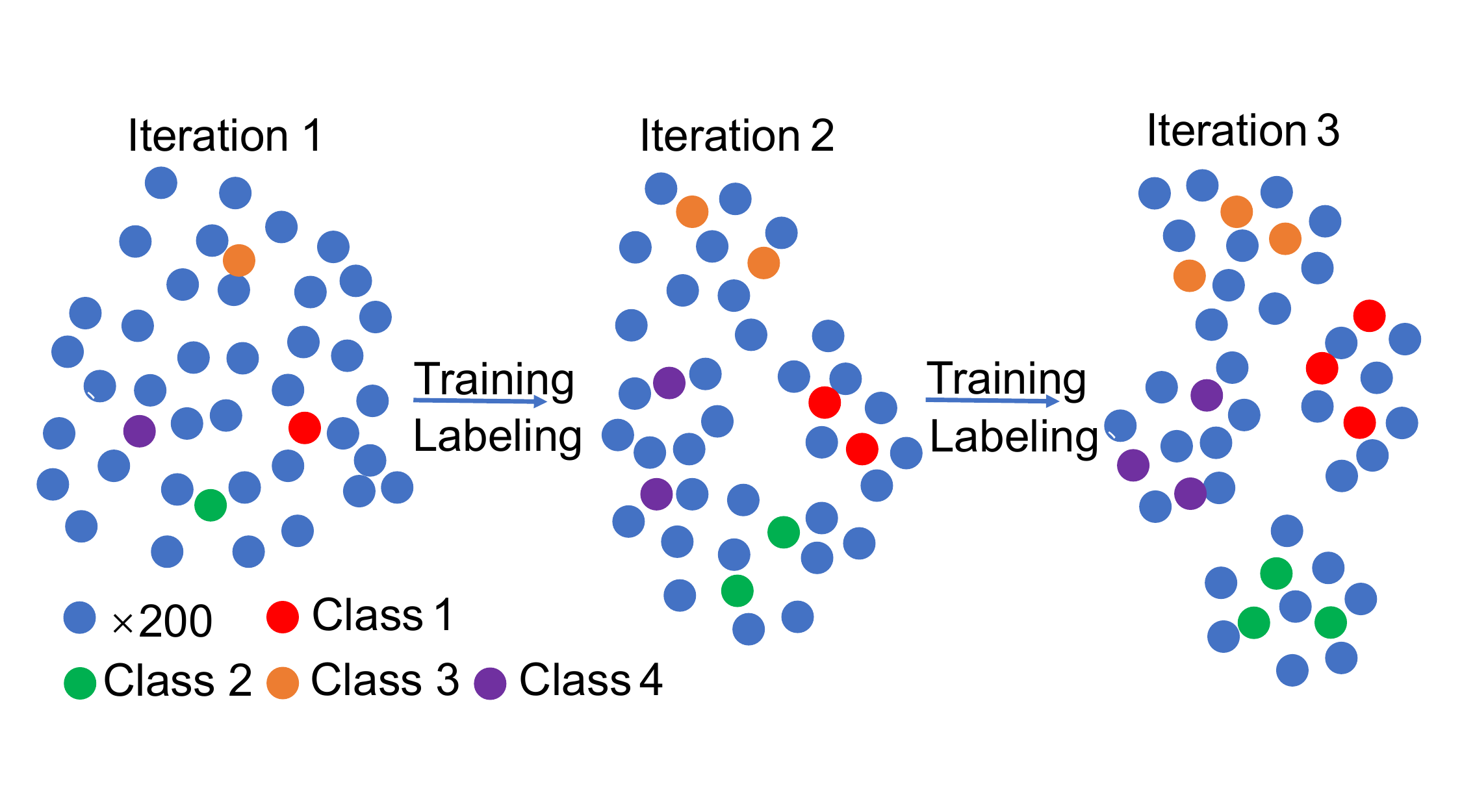}
    \end{minipage}
            \caption{Left: Iterate \& Cluster (IC) system structure. An active learning classifier $L_{CLASS}$ is added to the PC encoder-decoder network. Right: Three iterations of active selection of samples for annotation. In each iteration clusters in embedded space are being assigned and samples are chosen from each cluster (red, orange, green and purple) according to different active learning strategies. The process is repeated over several iterations.}
        \label{fig:selection}
\end{figure}


\textbf{\textit{Data processing}.} The dataset $\mathcal{X}=\{X\}$ of keypoints obtained from video frames consists of sequences (samples), where each sample $X$ is represented as $X = [x_1, x_2, ..., x_T]$, where $x_i$ is the keypoints position at time $i$,  $x_i \in R^{N \times D}$, here $N$ is the number of keypoints, $D$ is the dimension of the keypoints, $D=2, 3$. Our system is applicable to various types of keypoints and camera views from which the keypoints could be obtained.
For example, we demonstrate an action recognition of mouse movements from cage recordings~\cite{jhuang2010automated}, see Fig.~\ref{fig:Teaser}. For these videos, we identify 8 key points and track them with DeepLabCut \cite{mathis2018deeplabcut} to obtain a database of samples.
For the datasets with keypoints obtained from video frames recorded from multiple views (e.g. NW-UCLA, UWA3D) we follow the procedure of transforming them to a view invariant representation~\cite{su2019predict, shotton2011real}. 

\textbf{\textit{Unsupervised latent representation}.}\label{unsupervised} Previous work shows that RNN, in particular, an encoder-decoder sequence to sequence (Seq2Seq) networks, initialized with random weights, can self-organize their latent state space to cluster sequences in a low-dimensional embedding space~\cite{su2019predict,farrell2019recurrent}. Predict \& Cluster(PC) system which incorporates Seq2Seq type network was shown to use this property for action recognition. The network is trained to regenerate by the decoder the input sequence into the encoder, see Fig.~\ref{fig:selection}-(PC). The last state of the encoder can be used for classification of action using a simple classifier like KNN~\cite{su2019predict}. PC system is structured as follows. The encoder incorporates a bidirectional GRU cells which receive as input the values of the sequence. The forward last hidden state of the bi-directional encoder is denoted as $H_f$ while the backward part is denoted as $H_b$. These states are concatenated into $[H_f, H_b]$ and are passed to the decoder. 
With such unsupervised learning strategy PC achieved comparable performance to supervised methods when KNN classifier was set to classify the last latent state of the encoder $[H_f, H_b]$. This evaluation, however, relies on the strength of KNN classifier and assumes that it has all the labels in the training set. If only part of the labels is being used for the KNN classifier, then the performance is still admissible, however not as optimal as in the former case.

\textbf{\textit{Classification.}} We propose to boost the performance of such a system with a semi-supervised active learning strategy when annotations are unavailable to train the KNN classifier. We call our proposed system Iterate \& Cluster (IC) since we add a classifier to the last hidden state of the encoder, i.e., a classifier $L_{CLASS}$ to $[H_f, H_b]$, which iteratively selects samples for annotation, see Fig.~\ref{fig:selection}-IC.
IC could be trained with two different strategies:
\begin{itemize}[leftmargin=1.5em, topsep=0pt, itemsep=0pt, parsep=0pt]
\setlength\abovedisplayskip{0pt}
\item[(i)] The Seq2Seq is initialized randomly and trained along with the classifier. The loss in such training is a combined loss between classification loss $L_{CLASS}$ and prediction loss $L_{PRED}$ to learn essential features.
$L_{CLASS}$ in this case is Cross Entropy Loss, such that the output of the classifier is $y_{i}^{j}$, $y_{i}^{j} = 1$ indicating the sample $X_i$ belongs to class $j$ otherwise $y_{i}^{j}=0$. $L_{CLASS}$ is expressed as 
\setlength\abovedisplayskip{0pt}
\[
L_{CLASS} = \sum_{j=1}^{C}-y_{i}^{j}*\log(p_j(X_i)),
\]
\setlength\abovedisplayskip{0pt}
where $p_j(X_i)$ is the output probability of a sample $i$ to belong to class $j$, $C$ is the number of class. Such strategy turns out to be less effective since it does not use the self-organization feature of PC and is demonstrated for NW-UCLA dataset (named as IC) for comparison with methods that do leverage the PC representation for iterative label selection (IC-EP, IC-KR, IC-KT, IC-KEP). 
\setlength\abovedisplayskip{0pt}
\item[(ii)] The second strategy consists of two phases. In the first phase we train the PC network first (according to $L_{PRED}$) and then to use its self organized latent space to train the classifier further. The training for this later stage is according to a combined loss of $L_{IC} = 0.5L_{CLASS} + 0.5L_{PRED}$. This strategy turns out to be effective for enhancing the recognition accuracy. Furthermore, we observe that the accuracy depends on the samples being chosen. We therefore proceed and develop several selection strategies which leverages the representation achieved by PC training in the first phase to actively select samples for annotation that will increase the overall accuracy of the recognition task.
\end{itemize}

\textbf{\textit{Iterative Label Selection.}}
We define several selection methods for selecting samples for annotation and boosting the overall action recognition performance. The samples are selected in an incremental way (iterative) and then the network is trained on the current and previous selected samples with $L_{IC}$ loss.\\\\
\textit{Active Learning.} We follow a standard AL procedure~\cite{activ2012, gal2017deep} in which according to the output from the classifier layer we obtain the probability $p_i^c$ for sample $i$ to belong to its most possible class $c$, where for all samples this will correspond to an uncertainty vector $P = [p_1^c, ..., p_i^c, ...]$.  The main concept of AL strategies is to select the samples that the network is most uncertain about. The first approach (IC-PB) is to select $M$ samples corresponding to $M$ smallest values in $P$. These correspond to the samples that the classifier is most uncertain about. The second approach (IC-EP) is to compute the entropy of prediction for each sample $i$, i.e., 
\setlength\abovedisplayskip{0pt}
\[  H(X_i) = -\sum_{c=1}^{C}(p_i^clog(p_i^c)),~~~H = [H(X_1), ..., H(X_i), ...].
\]
\setlength\abovedisplayskip{0pt}

The IC-EP strategy selects $M$ samples with the maximal $H$ (entropy) values. In our experiments the IC-EP strategy turns out to be more successful than IC-PB and therefore results of IC-PB are included in the Supplementary materials. In addition to the samples for which we detect high uncertainty, we would also like to select the samples that are most representative for each class to have a balanced representation of the actions. Indeed, it is typically the case that the most uncertain samples are located in proximity to each other in the latent embedding space of the encoder and training on these can be misleading and not capturing balanced representation of the actions. We thereby propose another class of sampling methods that takes into account the clustering in the embedded space of the latent states of the encoder.\\\\
\textit{K Learning.} In this type of annotation selection we would like to obtain samples evenly distributed between classes of actions to represent the whole dataset properly. We therefore propose to perform a clustering of the encoder latent states space, i.e., cluster all samples $[H_b,H_f]$. We use $K-means$ for clustering, however other approaches, such Hierarchical clustering \cite{johnson1967hierarchical} could be applied as well in a similar fashion. For each cluster in this space and each sample, we compute the probability, $p_i^c$ for a sample $i$ to be located in its designated the cluster $c$. With this information we can select samples in several ways. For example,
a promising strategy is to select samples that are closest the clusters centers or alternatively it is possible to select according to the samples with smallest distances to cluster centers (IC-KT). Another possible selection method is to randomly select a fixed number of samples from each cluster (IC-KR). In practice, the number of samples selected in each iteration is at most 2C where C is the number of clusters, such that we choose a small number of samples and attempt to achieve maximal performance with each iteration. Since AL Learning and K Learning complement each other it is possible to combine the selection procedure which we describe next.
\\\\
\textit{KAL Learning.} To combine both types of sampling methods, we compute the entropy $H_c$ or the uncertainty $P_c$ for all samples in cluster $c$. Then for each cluster $c$, we select the samples according to their rank in $H_c$ (IC-KEP) or rank in $P_c$ (IC-KPB). This guarantees that within each cluster samples with most uncertainty are being selected early on, see comparison of the different strategies performance in Table~\ref{euclid} and the pseudo-code for the selection \textbf{Algorithm~\ref{al:algorithm}} below.
\begin{algorithm}
\caption{Iterative Label Selection}\label{euclid}
\begin{algorithmic}[1]
\Procedure{Label Selection}{}
\State $N \gets \textit{Number of class for current dataset}$
\State $per \gets \textit{Percentage of samples to choose from for each iteration}$
\State $Niter \gets \textit{number of iterations to choosing samples}$
\State $Strategy \gets \textit{Choosing method: \textbf{EP}, \textbf{PB}, \textbf{KEP}, \textbf{KPB},\textbf{KT}, \textbf{KR} }$
\State $\textit{M} \gets \textit{Number of clusters to form using Kmeans}$
\While{$Niter > 0 $}
\State $S_m \gets \text{Compute} \textit{samples in each cluster }$
\State $mtric \gets \textit{Compute metric for each sample in each cluster according to Strategy}$
\State $Nepoches \gets \textit{Number of epochs to train new added samples}$
\For {$i = 1:M$}
\State $Nsamples \gets \text{per}*\textit{Number of samples in the cluster}$
\State $NewSample \gets \textit{Strategy(metric, Nsamples)}$
\State $Sample \gets \textit{Sample + NewSample}$
\EndFor
\State $Niter \gets Niter-1$
\State \textbf{Annotate} \textit{the selected labels}
\State \textbf{Train} \textit{the network for $Nepoches$}
\EndWhile
\EndProcedure
\label{algorithm:algorithm}
\end{algorithmic}
\label{al:algorithm}
\end{algorithm}

\section{Results \& Evaluations}
\textbf{\textit{Datasets.}}
We evaluated the performance of our system (IC) on action recognition in three benchmark datasets. Two benhmarks are for human body skeleton-based action recognition, datasets \textit{North-Western UCLA (NW-UCLA)}~\cite{wang2014cross} and~\textit{Multiview Activity II (UWA3D)}~\cite{rahmani2014hopc}. We also consider a dataset of mouse actions, \textit{Home-cage Behavior Mouse} dataset~\cite{jhuang2010automated}.\\\\ \textbf{\textit{NW-UCLA}} dataset is captured by RGB three Kinect Automated Home-Cage Behavioral cameras containing depth and human skeleton data from three different views. The dataset includes 10 different action categories performed by 10 different subjects repeating from 1 to 10 times. We use the first two views as a training set, the last view (V3) as a test set following the same procedure as in~\cite{su2019predict, liu2017enhanced, wang2014cross}. \\\\
\textbf{\textit{UWA3D}} contains 30 human actions categories. Each action is performed 4 times by 10 subjects recorded from four views frontal, left and right sides, and top respectively. There are 15 key points for each action. We selected the first two views as training sets and the third view as the test set (expected to be more challenging in perfromance of related work \cite{su2019predict, zhang2017view}). Results for the fourth view are included in the Supplementary materials. View-invariant transformation is applied for these two datasets prior to analysis. In particular, keypoints representing human body joints located in the view variant space represented as $X^V$ would be transformed such that that connection between left hip and right hip is parallel to the ground, connection between spine and root is perpendicular to ground, etc. This is a similar procedure as in~\cite{su2019predict, shotton2011real}.\\\\
\textit{\textbf{Mouse}} To evaluate the generalization of IC to different types of keypoints, such as animal action and also to test few-shot active learning strategy (considering very small number of annotations) we consider the Automated Home-Cage Behavioral dataset. The dataset includes 8 different mouse behavioral phenotypes recorded using one camera in front of the cage. To extract the keypoints of the mouse we use DeepLabCut~\cite{mathis2018deeplabcut} to label 8 key points (snout, left-forelimb, right-forelimb, left-hindlimb, right-hindlimb, fore-body, hind-Body, and tail) and acquire the positions of the keypoints, as well as the confidence for prediction of their position. To make sure the information provided by DeepLabCut is reliable, we define a probability threshold and only the sequences that pass that threshold will be used. We further describe the selection and threshold process in the Supplementary material. 
There is a total of 1504 final samples in the dataset and we split them to training  and testing sets as split of $70\%$/$30\%$ respectively.\\

\begin{figure}[H]
    \centering
	\includegraphics[scale = 0.42]{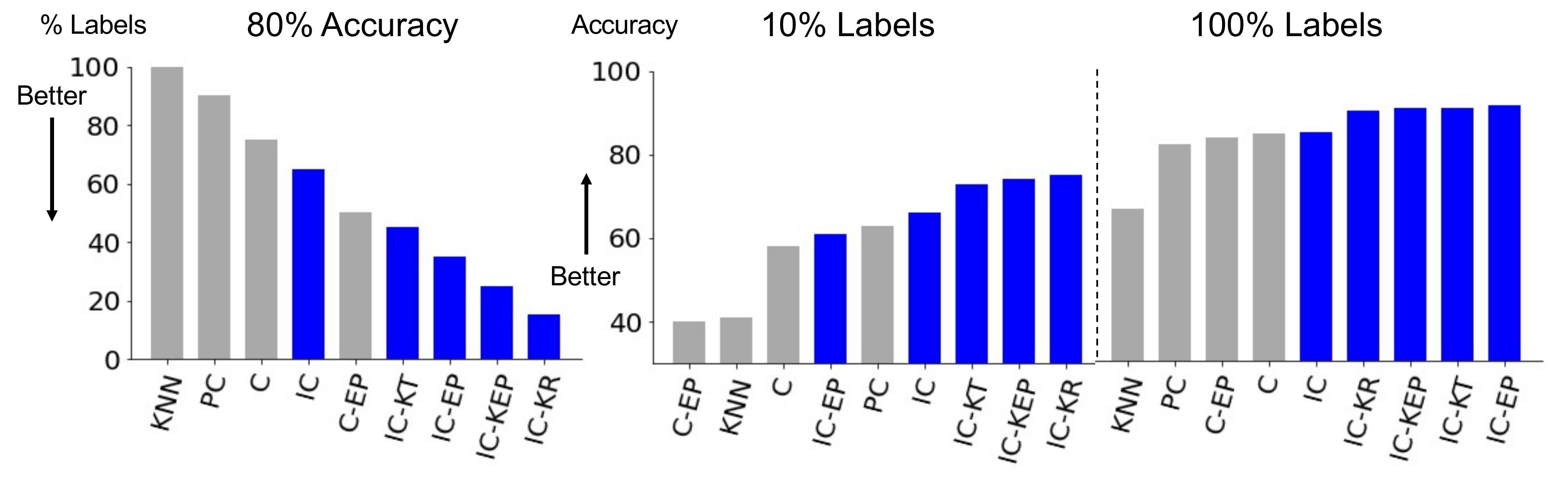}
	\caption{Left: Number of annotated labels (in $\%$) required to acheive 80\% accuracy on NW-UCLA dataset benchmark. Middle: The accuracy (in $\%$) each method achieves with the annotation of $10\%$ of the labels in the dataset. Right: The accuracy (in $\%$) each method achieves with annotation of $100\%$ of the labels in the dataset.} 
	\label{fig:10ucla}
\end{figure}

\begin{figure}[H]
	\includegraphics[scale = 0.63]{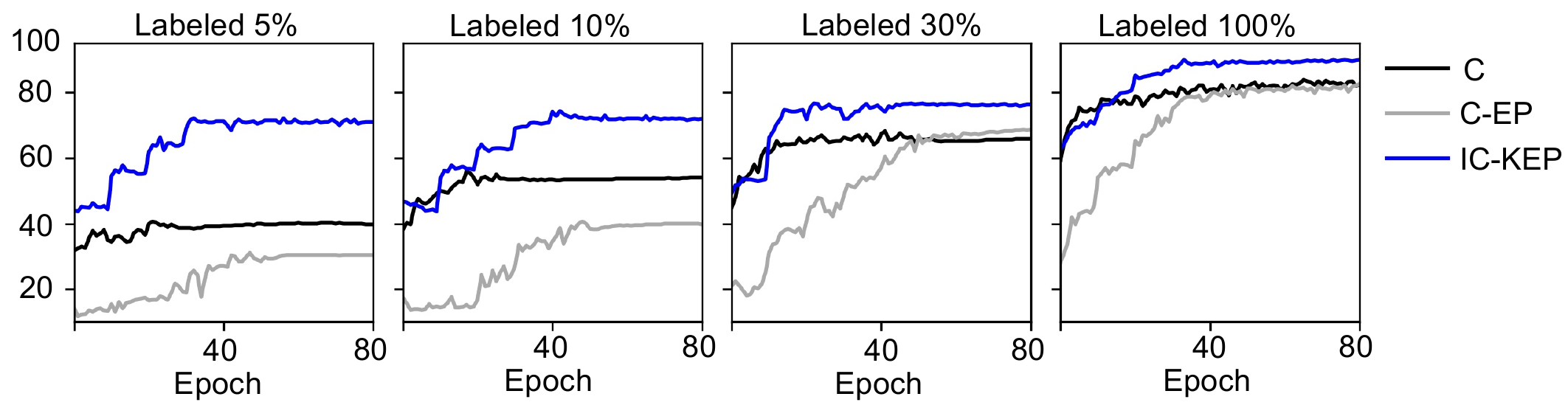}
	\caption{Training trajectory usind different number of labels $5\%, 10\%, 30\%, 100\%$ with three different methods}
	\label{fig:UCLA}
\end{figure}

\textbf{\textit{Implementation Details.}}
The unsupervised system (PC) is set to have an encoder as a 3 layer bidirectional GRU with 1024 hidden units for each layer. Outputs from last layer of the encoder are concatenated into a 2048 dimension vector, $[H_f,H_b]$ treating it as the input to the decoder. The decoder is a unidirectional one layer GRU followed by a linear regression layer with output dimension to be the same as input feature dimension predicting keypoints positions at each time step. Fixed-Weight PC training strategy was used. The action recognition task for Mouse dataset is easier than other tasks and therefore we use the same architecture but with fewer units: 1 encoder layer with 125 hidden units for each direction and the decoder dimension set to be 250 to fit the encoder hidden output. A fully connected neural network activated with Softmax is added to the last hidden state of the encoder (Fig.~\ref{fig:selection}). We use Adam optimizer and initialize the learning rate to be 0.0001. Learning rate decreases by $5\%$ every 50 epochs. 
\begin{table*}
\vspace{0pt}
\centering
\begin{tabular}{@{}rrrrcrrrcrrr@{}}\toprule
& \multicolumn{5}{c}{\textbf{NW-UCLA}} \\
\midrule
\# Labels (\%) & $5$ & $10$ & $20$ &$50$ &$100$  \\ \midrule
C-EP &31.1&40.1&56.3&\textbf{80.9}&83.5\\
KNN &36.1&39.8&47.4&55.7&66.5\\
C &40.7&55.9&58.3&71.5&\textbf{84.4}&\\
PC~\cite{su2019predict} &\textbf{52.0}&\textbf{65.2}&\textbf{66.1}&72.6&81.7\\
\midrule
\textcolor{blue}{IC (Our)} &64.1&66.3&66.3&79.8&84.8\\
\textcolor{blue}{IC-EP (Our)}&59.8&61.0&75.0&\textbf{85.7}&\textbf{91.3 }&\\
\textcolor{blue}{IC-KR (Our)} &70.4 &\textbf{75.4 }&\textbf{81.7 }&84.8&90.0 \\
\textcolor{blue}{IC-KT (Our)} &64.6&72.8 &75.2 &80.9&90.4 &\\
\textcolor{blue}{IC-KEP (Our)} &\textbf{72.2}&74.4&78.5&84.6&90.4 &\\
\bottomrule
\end{tabular}
\caption{Comparison of different approaches on NW-UCLA dataset. Previous or standard methods in black. IC methods in blue.}
\label{resultucla}
\end{table*}\\
\\\textbf{\textit{Evaluation.}}
We test on NW-UCLA dataset to evaluate how accuracy and the number of annotations vary for different IC system variants (5 variants, marked in blue) and in comparison to other methods (4 methods : KNN trained on the provided samples only, PC, C (classification with $L_{CLASS}$ only) and C-EP (classifier with active EP active learning) marked in gray)(Fig.~\ref{fig:10ucla}). In particular, we compare $(i)$ the number of annotated labels each method would require to achieve 80\% accuracy and $(ii)$ the accuracy of each method given 10\% or 100\% of the annotations that can be obtained.  We find that to achieve $80\%$ accuracy, IC methods require significantly less samples than other methods. \textit{IC-KR} (IC with random selection from clusters) uses the least number of samples (15\%). The plain \textit{IC} which does not leverage the embedding requires 65\% of labeled samples and is similar to the performance of \textit{C-EP} which requires 50\% of labels to achieve that accuracy. This shows that the use of latent space embedding for selection is critical. The KNN method achieves only 66.5\% on full 100\% of the data and never reaches 80\% accuracy. When we fix certain number of labels to be annotated, i.e., $10\%$ in middle of Fig.~\ref{fig:10ucla}, we find that four of our methods achieve higher accuracy than the compared methods. \textit{IC-EP} is the only method that is not preferable than the compared \textit{PC} which again indicates the importance of the unsupervised latent state embedding which more optimally directs the performance of active learning. The best performance achieved with 10\% annotation of the data is 75.4\% and is with \textit{IC-KR}. When using all labels ($100\%$) our methods are able to achieve top performance outperforming \textit{C} with the same network structure which indicates that the intial an iterative selection of labels boosts the overall performance as well. We also compare the performances with $5\%$, $20\%$, and $50\%$ of annotations in Table~\ref{resultucla} for NW-UCLA dataset. We mark the best performance of compared methods and our methods with bold. For small number of annotations (i.e 5\%) we observe a significant boost of more than 20\% in accuracy from 52\% to 72.2\% for the combined active learning and embedded clusters strategy (\textit{IC-KEP}). As more annotations are acquired the advantage of the IC methods approach is kept, though we see that different strategies may be more effective than other in specific number of annotations. In Fig.~\ref{fig:UCLA} we elaborate on that and show how iterative addition of labels affects accuracy for \textit{C} (standard classification), \textit{C-EP} (EP active learning) and \textit{IC-KEP} (Our). It can be observed that there are rapid improvements in \textit{IC-KEP} each time samples are annotated and also in general that \textit{IC-KEP} opens a gap early on with other the two methods especially when the number of annotations allowed is small (i.e. 5\%,20\%). The plot indicating PC initialization and KAL matters for achieve satisfying performance either we have a few labeled samples or we have full label.\\
From all the methods tested on NW-UCLA, we select three of our methods of \textit{IC-KR}, \textit{IC-KT}, \textit{IC-KEP} which achieve top performance for limited number of samples and compare with four methods \textit{KNN}, \textit{C-EP}, \textit{C} and \textit{PC} on UWA3D dataset  and and also on Mouse dataset in Table~\ref{table:table2}. On UW3AD, our methods outperform all other methods for different number of labels and increase in performance as more annotations are provided is similar to \textit{NW-UCLA} dataset.
\\Also, we test how our methods perform when the number of labels is limited on the Mouse dataset. We can find \textit{IC-KR} method works best when we have only $5\%$, $10\%$ and $20\%$ labeled samples. \textit{IC-KEP} outperforms others with $50\%$ labels. Our methods beat other methods with limited number of labels. When we have full labels the proposed methods works similar to traditional actively learning methods \textit{C-EP}. Best of our methods, \textit{IC-KEP} achieves $95.3\%$ while \textit{C-EP} achieves $95.8\%$. One possible explanation for why \textit{C-EP} does not work when number of labeled samples are few, will be that \textit{C-EP} selection really depends on how samples are distributed in embedding space, sometimes it may achieve higher accuracy when selected samples are representative in whole embedding space, but much worse when those samples are cluttered in one location. The more samples going to be selected for annotations the more likely those selected samples are representative. The methods we proposed overcome this defect.
\setlength\abovedisplayskip{0pt}
\begin{table}
\centering
\begin{tabular}{@{}rccccc||cccc|c@{}}\toprule
& \multicolumn{5}{c}{\textbf{UWA3D}} & \multicolumn{5}{c}{\textbf{Mouse}} \\
\midrule
\# Labels (\%) & $5$ & $10$ & $20$ &$50$ &$100$ & $5$ & $10$ & $20$ &$50$ &$100$  \\ \midrule
KNN & 18.7&20.3&24.0&32.9&40.2 &58.4&66.7&72.7&75.1&77.3\\
C-EP &\textbf{21.9}&27.6 &35.8&47.2&56.9  
&70.7&78.9&74.9&91.3&\textbf{95.8}\\
C &18.7&22.4&30.9&45.1 &55.3  
&\textbf{84.3}&\textbf{86.7}&\textbf{91.6}&92.2&93.4 \\
PC\cite{su2019predict}&20.7&\textbf{32.1}&\textbf{46.3 }&\textbf{54.1}&\textbf{60.2} 
&82.7&86.3&90.2&\textbf{93.6}&94.5\\
\midrule
\textcolor{blue}{IC-KR(Our)}&26.0 &\textbf{39.4 }&47.6&61.4&64.6   
&\textbf{89.1}&\textbf{90.7}&\textbf{93.4}&93.6&93.8\\
\textcolor{blue}{IC-KT(Our)}&24.4&36.2&\textbf{51.6 }&\textbf{68.5}&\textbf{67.5}
&68.3&83.8&86.3&92.0&93.4\\
\textcolor{blue}{IC-KEP(Our)}&\textbf{28.9}&38.6 &48.4&60.6&65.9 &69.0&85.4&86.5&\textbf{94.7}&\textbf{95.3}\\ 
\bottomrule
\end{tabular}
\caption{Comparison between our methods IC-KR, IC-KT, IC-KEP with C-EP, C, PC\cite{su2019predict} in UWA3D and Mouse dataset}
\label{table:table2}
\end{table}
\setlength\abovedisplayskip{0pt}


\section{Conclusion}
We proposed a semi-supervised active learning framework, \textit{Iterate \& Cluster}, for action recognition with a limited number of annotations. We have shown that \textit{(i)} our system can effectively enhance action recognition even when only 5\% of the dataset is being queried. \textit{(ii)} Our key idea is to use an unsupervised approach to initialize our IC system and guide active learning selection. \textit{(iii)} Our simple, general and effective training procedure on the full annotated dataset is on par or outperforms state-of-the-art supervised learning approaches.

\bibliographystyle{unsrt}

\end{document}